\documentclass[sigconf]{acmart}
\usepackage{colortbl}
\usepackage{graphicx}
\usepackage{subfig}
\usepackage{multirow}
\usepackage{tabulary}
\usepackage{pifont}

\usepackage{algorithm}
\usepackage{algorithmicx}%
\usepackage{algpseudocode}%

\usepackage{listings}

\AtBeginDocument{%
  \providecommand\BibTeX{{%
    \normalfont B\kern-0.5em{\scshape i\kern-0.25em b}\kern-0.8em\TeX}}}

\setcopyright{acmlicensed}
\acmDOI{3690624.3709266}

\acmConference[KDD '25]{the 31th ACM SIGKDD
Conference on Knowledge Discovery and Data Mining V.1}{August 3--7,
  2025}{Toronto, ON, Canada}
\begin{document}

\title{TSINR: Capturing Temporal Continuity via Implicit Neural Representations for Time Series Anomaly Detection}


\author{Mengxuan Li}
\orcid{0001-7278-7891}
\affiliation{%
  \institution{Zhejiang University}
  \city{Hangzhou}
  \country{China}
}
\email{mengxuanli@intl.zju.edu.cn}

\author{Ke Liu}
\affiliation{%
  \institution{Zhejiang University}
  \city{Hangzhou}
  \country{China}
  }
\email{keliu99@zju.edu.cn}

\author{Hongyang Chen}
\affiliation{%
 \institution{Zhejiang Lab}
 \city{Hangzhou}
 \country{China}
 }
\email{hongyang@zhejianglab.com}

\author{Jiajun Bu}
\affiliation{%
  \institution{Zhejiang University}
  \city{Hangzhou}
  \country{China}}
\email{bjj@zju.edu.cn}

\author{Hongwei Wang}
\authornotemark[1]
\affiliation{%
 \institution{Zhejiang University}
 \city{Hangzhou}
 \country{China}}
\email{hongweiwang@intl.zju.edu.cn}

\author{Haishuai Wang}
\authornotemark[1]
\affiliation{%
  \institution{Zhejiang University}
  \city{Hangzhou}
  \country{China}
}
\email{haishuai.wang@zju.edu.cn}
\thanks{*Corresponding authors.}



\begin{abstract}
Time series anomaly detection aims to identify unusual patterns in data or deviations from systems’ expected behavior. The re-construction-based methods are the mainstream in this task, which learn point-wise representation via unsupervised learning. However, the unlabeled anomaly points in training data may cause these reconstruction-based methods to learn and reconstruct anomalous data, resulting in the challenge of capturing normal patterns. In this paper, we propose a time series anomaly detection method based on implicit neural representation (INR) reconstruction, named TSINR, to address this challenge. Due to the property of spectral bias, TSINR enables prioritizing low-frequency signals and exhibiting poorer performance on high-frequency abnormal data. Specifically, we adopt INR to parameterize time series data as a continuous function and employ a transformer-based architecture to predict the INR of given data. As a result, the proposed TSINR method achieves the advantage of capturing the temporal continuity and thus is more sensitive to discontinuous anomaly data. In addition, we further design a novel form of INR continuous function to learn inter- and intra-channel information, and leverage a pre-trained large language model to amplify the intense fluctuations in anomalies. Extensive experiments demonstrate that TSINR achieves superior overall performance on both univariate and multivariate time series anomaly detection benchmarks compared to other state-of-the-art reconstruction-based methods. Our codes are available \textcolor{blue}{\href{https://github.com/Leanna97/TSINR}{here}}.
\end{abstract}

\ccsdesc[500]{Computing methodologies~Neural networks}
\ccsdesc[500]{Mathematics of computing~Time series analysis}

\keywords{time series anomaly detection, implicit neural representations, unsupervised learning}


\maketitle

\section{Introduction}
Time series anomaly detection, which aims to identify unusual patterns or events across a sequence of data points collected over time \cite{li2023deep}, has attracted a lot of attention recently. In many fields (e.g., finance, healthcare, manufacturing, and fault diagnosis), monitoring time-varying data to identify anomalies is crucial for detecting unusual behavior, potential issues, or security threats \cite{wang2018learning,jie2024disentangled,wang2019time,wu2024spatio,10172249}. For example, in the finance field, detecting anomalous behavior in credit card data allows the prevention of theft or fraudulent transactions committed by an unauthorized party \cite{singh2012outlier}. In industrial processes, identifying anomalies helps secure safe operations, averting safety concerns and mitigating economic losses \cite{10257663}. 

One of the major challenges for anomaly detection lies in anomalies may be rare, subtle, or have different shapes, requiring sophisticated algorithms to distinguish them from normal patterns. Moreover, time series typically exhibit trends, seasonality, and temporal dependencies, making it challenging to model such complex features. Since anomalies are typically rare and new anomalies may arise, it is difficult or expensive to collect a sufficient amount of labeled data. As one of the unsupervised-based methods, reconstruction-based methods tackle this problem by reconstructing data to learn point-wise feature representations to uncover normal patterns in the data \cite{zhong2024refining,zhang2023stad}. The reconstruction error, i.e., the difference between the input data and its reconstructed version, serves as a natural anomaly score. The points that have high anomaly scores are considered as anomalies, which makes it easy to interpret and understand the results.

\begin{figure*}[]
\centering
\subfloat[]
{
 \centering
 \includegraphics[width=8.3cm]{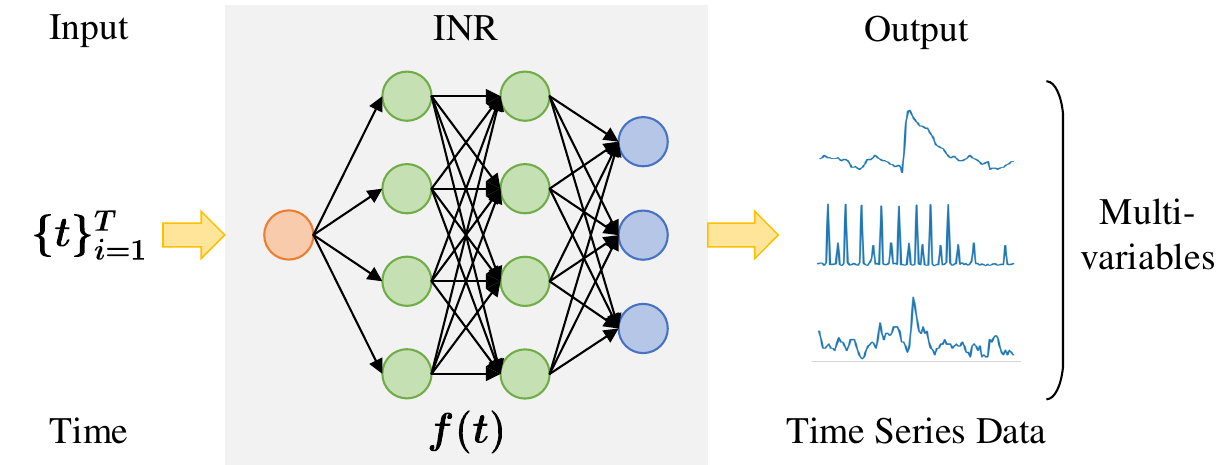}
 \label{fig_motivation1}
}
\hfill
\subfloat[]
{
 \centering
 \includegraphics[width=8.3cm]{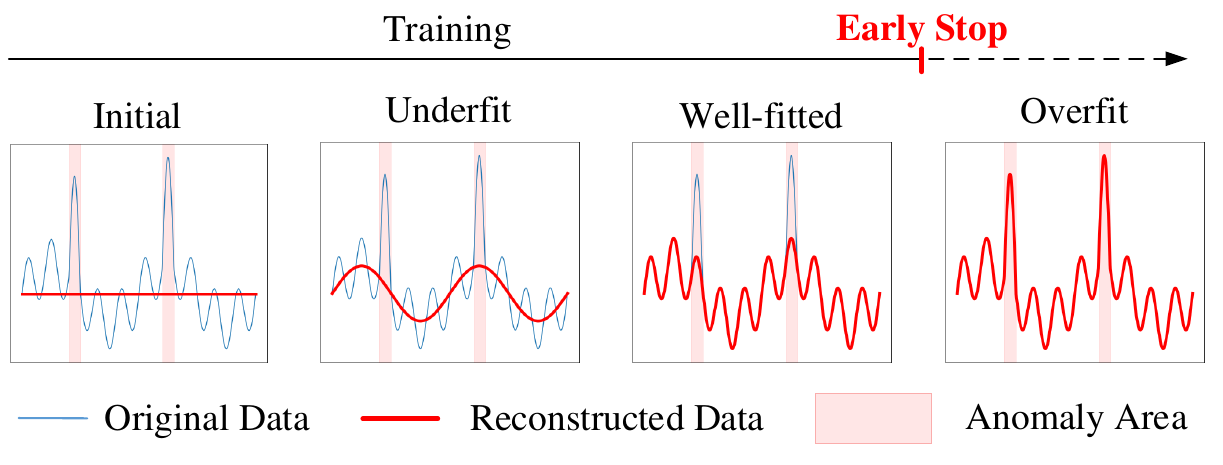}
 \label{fig_motivation2}
}
\vspace{-0.3cm}
\caption{(a) The diagram of INR for time series data. (b) The spectral bias property of INR to prioritize the low-frequency signals is advantageous for accomplishing time series data anomaly detection tasks.} 
\label{fig_motivation}
\vspace{-1.1em}
\end{figure*}

However, it is challenging to distinguish normal and anomaly patterns because normal and anomalous points may coexist within a single instance, and anomalies may occur in the unlabeled training data \cite{yang2023dcdetector}. In addition, time series data often contains intricate patterns, and anomalies might exhibit subtle deviations from normal ones. Therefore, models may be forced to learn and reconstruct anomalous data, this makes learning a reconstruction model that can effectively capture the normal pattern challenging. This issue is also pointed out in previous work \cite{yang2023dcdetector}.
Recently, implicit neural representation (INR) has become a powerful tool for continuous encoding of various signals by fitting continuous functions \cite{liu2023partition,molaei2023implicit}. Figure \ref{fig_motivation1} depicts the diagram of INR within the context of time series data. As a continuous function, INR captures the temporal continuity of the time series, where the input is a timestamp, and the output is the corresponding value of this timestamp. In line with existing reconstruction-based anomaly detection methods, INR is also learned through a reconstruction task, making it inherently feasible for time series anomaly detection. In addition, INR possesses a spectral bias property, enabling it to prioritize the learning of low-frequency signals. Most efforts aim to alleviate this property to enhance the fitting capability for high-frequency signals \cite{sitzmann2020implicit,tancik2020fourier}. Conversely, this property is advantageous for accomplishing time series anomaly detection. As in Figure \ref{fig_motivation2}, the spectral bias property enables INR to prioritize the smooth normal points and exhibit poorer performance for high-frequency abnormal data. Therefore, we aim to leverage the ability of INR to capture continuous representations and its sensitivity to anomalous data, thereby addressing the challenges of existing reconstruction-based methods. 

In this paper, we propose a time series anomaly detection method based on INR reconstruction (TSINR for short). Specifically, we introduce a transformer-based architecture to predict the INR parameters of the given time series data.  To better learn and reconstruct time series data, we design a novel form of continuous function to decompose time series \cite{cleveland1990stl,fons2022hypertime}. The designed function mainly comprises three components and individually learns the trend, seasonal, and residual information of time series. In addition, to further enhance the capability of INR to capture inter- and intra-channel information, we propose a group-based architecture to specifically learn the complex residual information. Simultaneously, we leverage a pre-trained large language model (LLM) to encode the original data to the feature domain. This encoding process particularly amplifies the fluctuations of anomalies across both the time and channel dimensions. By doing so, we enable TSINR to more effectively distinguish between normal and abnormal data points. The major contributions of this paper are summarized as follows:
\vspace{-1em}
\begin{itemize}
    \item We utilize the spectral bias property of INR to prioritize fitting low-frequency signals and enhance sensitivity to discontinuous anomalies, thereby improving anomaly detection performance. A transformer-based architecture is employed to generate the parameters for INR, requiring only a single forward step in the inference phase.
    
    \item We design a novel form of INR continuous function, which mainly consists of three components to implicitly learn the unique trend, seasonal, and residual information of time series. Furthermore, a group-based strategy is proposed to further learn intricate residual information.

    \item We leverage a pre-trained LLM to encode the original time series to the feature domain, enabling amplification of the fluctuations of anomalies in both time and channel domains that facilitate INR to be further sensitive for noncontinuous anomaly areas. Ablation studies and visual analysis validate the aforementioned capacity to better distinguish anomaly points via our proposed framework. 

    \item Extensive experiments demonstrate the overall effectiveness of TSINR compared with other state-of-the-art methods on seven multivariate and one univariate time series anomaly detection benchmark datasets. 
\end{itemize}

\begin{figure*}[]
\centering
\includegraphics[width = 0.87\textwidth]{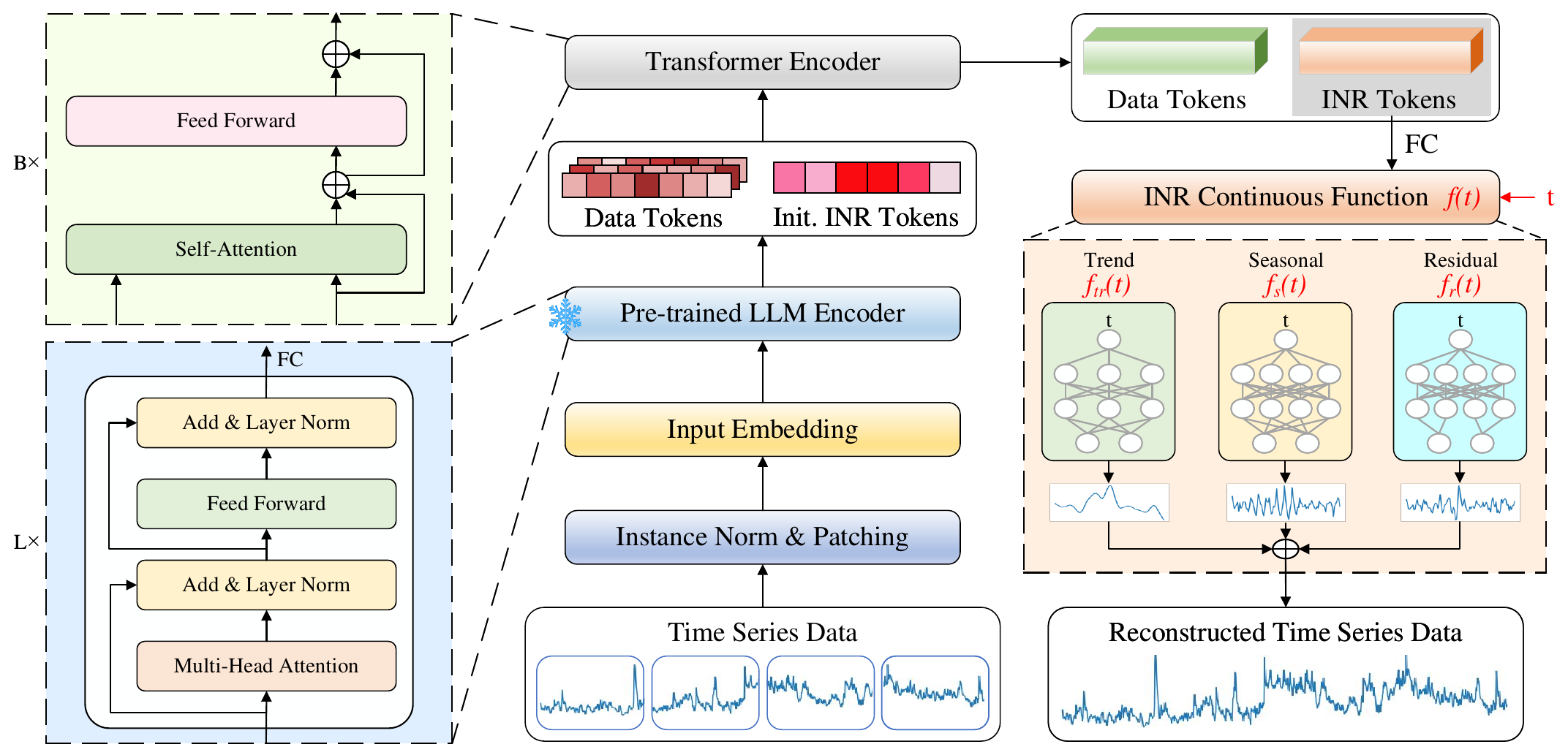}
\vspace{-0.1cm}
\caption{The overall workflow of the proposed TSINR method. The INR tokens predicted by the transformer encoder are the parameters of the INR continuous function. And the input of the INR continuous function is the timestamp $t$.}
\label{fig_workflow}
\vspace{-1em}
\end{figure*}

\vspace{-0.5em}
\section{Related Work}
\subsection{Time Series Anomaly Detection}
Time series anomaly detection methods primarily include statistical, classic machine learning, and deep learning methods. Statistical methods rely on analyzing the statistical properties of the data to identify patterns that deviate from the expected behavior. They are valuable for their simplicity and interpretability, but have limitations in capturing complex patterns \cite{cleveland1990stl, jie2024disentangled}.


Classic machine learning methods rely on manual feature extraction and various algorithms like clustering \cite{ruff2018deepsvdd,shin2020itad}, density estimation \cite{breunig2000lof,yairi2017data-MPPCACD,zong2018deep-DAGMM}, and isolation forests \cite{liu2008isolation-iForest} to identify anomalies in structured data. However, because they require manual feature extraction and selection, they can be labor-intensive and less effective at capturing complex patterns in data.


Deep learning methods automatically learn the features of data through deep neural networks without the need for manual intervention, and are adept at handling high-dimensional, unstructured data. They can be broadly categorized into supervised and unsupervised learning algorithms. Supervised methods are trained with labels to learn and classify both normal and anomalous behavior in the given time series data, such as NFAD \cite{ryzhikov2021nfad}
and MultiHMM \cite{li2017multivariate}. However, annotating data is challenging due to the rarity of anomalies and the emergence of new anomalies. This makes it difficult to achieve effective labeling, leading to limitations in the performance of supervised methods in detecting anomalies.

On the contrary, unsupervised methods distinguish anomalous points from normal ones without relying on prior knowledge. The unsupervised methods mainly comprise 2 categories: forecasting-based methods and re-construction-based methods. Forecasting-based methods train a model to predict future values based on past observations and then identify anomalies by comparing the actual values with the predicted ones, such as ARIMA \cite{yaacob2010arima} and Telemanom \cite{hundman2018detecting}. Reconstruction-based methods learn and reconstruct the input data and then identify anomalies based on the difference between the original and reconstructed data, such as LSTM-VAE \cite{park2018multimodal}, BeatGAN \cite{zhou2019beatgan}, OmniAnomaly \cite{su2019robust}, and TranAD \cite{tuli2022tranad}.

Recently, several methods have been proposed to establish a universal framework that can effectively address a wide range of time series data tasks, such as FPT \cite{zhou2023one} and TimesNet \cite{wu2022timesnet}. Among these works, models designed for prediction tasks also perform well in anomaly detection tasks, such as DLinear \cite{zeng2023transformers}, ETSformer \cite{woo2022etsformer} and LightTS \cite{zhang2022less}. These methods demonstrate outstanding performance in learning time series features, thereby it makes sense to use these methods as baselines.

\vspace{-1em}
\subsection{Implicit Neural Representations}
Presently, INR \cite{xie2022neural} stands as a scorching topic in the domain of deep learning. It aims to learn a continuous function, often embodied as a neural network, for data representation. In this function, the input comprises coordinates, while the output consists of corresponding data values. INR learns continuous representations and has been widely applied in numerous scenarios, such as 2D image generation \cite{qin2022hilbert,strumpler2022implicit,zhou2021distilling} and 3D scene reconstruction \cite{ran2023neurar,kohli2020semantic,guo2022neural}, physics-informed problems \cite{raissi2019physics,pfrommer2021contactnets} 
and video representation \cite{chen2022videoinr,lu2023learning,mai2022motion}. 

Current methods for learning INR parameters predominantly rely on two main approaches: meta-learning \cite{liu2023partition} and feed-forward networks \cite{chen2022transformers, zhang2024attention-eccv}. The primary distinction between these approaches lies in how they handle test data. Meta-learning techniques are designed to quickly adapt to unseen test data by requiring only a few training steps or fine-tuning iterations for each new input. This allows for efficient generalization to novel tasks with minimal data. In contrast, feed-forward networks directly generate predictions in a single forward pass, leveraging pre-trained parameters without the need for task-specific adaptation during inference. In this paper, we use a transformer-based architecture to generate INR parameters and it requires only a single forward in the inference phase \cite{chen2022transformers}.

In addition, there exists a phenomenon known as spectral bias \cite{rahaman2019spectral}, where INR tends to prefer fitting the low-frequency components of the signal \cite{liu2023partition}. 
Since this characteristic can affect the ability of INR to model high-frequency data, most efforts are directed towards mitigating this effect \cite{sitzmann2020implicit,tancik2020fourier,liu2023implicit-icme}. In contrast, in time series anomaly detection, this property turns out to be advantageous. In time series data, normal points exhibit relative smoothness, whereas anomalous points possess strong discontinuity. Hence, we leverage this property of INR to prioritize fitting normal data with low frequencies, making it more sensitive to anomalous data.

\vspace{-1em}
\subsection{Implicit Neural Representations on Time Series Data}
Currently, there are some studies discussing the possibility of utilizing INR for time series representation. HyperTime \cite{fons2022hypertime} leverages INR to learn
a compressed latent representation for time series imputation and generation. TimeFlow \cite{naour2023timeflow} uses INR to capture continuous information for time series imputation and forecasting. 
In addition, the potential of employing INR for time series anomaly detection has not been fully explored yet. Only INRAD \cite{jeong2022inrad} attempts to adopt INR to represent and reconstruct time series data to identify anomalies. INRAD aims to utilize INR to overcome deep learning limitations, like complex computations and excessive hyperparameters (e.g., sliding windows). However, it requires training an INR network for each unseen time series data in test set, leading to additional training time and inefficiency in practical applications.

Different from INRAD, our method uses the spectral bias property of INR to mitigate the impact of unlabeled anomalies for the reconstruction model. With the Transformer integration, it can detect anomalies in unseen test data without retraining, enhancing efficiency for practical applications. In addition, compared to INRAD and other INR methods, our approach incorporates several specialized designs for time series anomaly detection. Firstly, we devise a novel form of INR continuous function to capture trend, seasonal, and residual information to address unique temporal patterns. Secondly, to handle multivariate time series, we introduce a group-based architecture to bolster the representational capacity of INR. Lastly, we leverage LLM to enhance anomaly detection by amplifying anomaly fluctuations, thereby boosting the sensitivity of INR to anomalies.


\vspace{-0.5em}
\section{Methodology}
In this paper, we propose TSINR, a novel time series anomaly detection method based on INR reconstruction. The core idea is to leverage the spectral bias phenomenon of INR to prioritize fitting smooth normal points, thereby enhancing sensitivity to discontinuous anomalous points. In this section, we present the problem statement and introduce the overall architecture, followed by the form of INR continuous function designed for time series data and the frozen pre-trained LLM encoder applied to amplify the fluctuations of anomalies from both time and channel dimensions. The anomaly criterion is demonstrated finally.

\subsection{Problem Statement}
Consider a time series $X$ with $T$ timestamps: $X=\left(x_1, x_2, \cdots, x_T\right)$,
where $x_t \in \mathbb{R}^d $ is the data point observed at a certain timestamp $t$ ($t \in \{1,2, \ldots, T\}$) and $d$ denotes the number of the data variables (\emph{i.e.}, data dimensionality). For a multivariate data, $d>1$. And for an univariate case, $d=1$. Given unlabeled input time series data $X_{train}$, for any unknown time series data $X_{test}$ with the same data dimensionality $d$ as $X_{train}$, we aim to predict $y_{test}=\left(y_1, y_2, \cdots, y_{T^{\prime}}\right)$, where $T^{\prime}$ is the length of $X_{test}$. And $y_{t^{\prime}} \in \{0,1\}$ denotes whether the data point is normal ($y_{t^{\prime}}=0$) or abnormal ($y_{t^{\prime}}=1$) at the certain timestamp $t^{\prime}$ ($t^{\prime} \in \{1,2, \ldots, T^{\prime}\}$).

\subsection{Overall Workflow}
Figure \ref{fig_workflow} shows the overall workflow of the proposed TSINR method. We employ a feed-forward transformer-based architecture to directly predict the whole weights of the INR of the given time series data \cite{chen2022transformers}. Unlike meta-learning based on gradient descent \cite{liu2023partition}, our method requires only a single forward step in the inference phase. Following a strategy similar to other transformer-based methods \cite{dosovitskiy2020image,zeng2023transformers}, the input time series data is normalized and segmented into patches. A frozen pre-trained LLM encoder is applied to map the input data into the feature domain to amplify the fluctuations of anomalies. Then the obtained features are converted to data tokens using a fully connected (FC) layer. Simultaneously, we initialize the corresponding INR tokens, which are learnable vector parameters. These data tokens and initialized INR tokens are fed together into a transformer encoder, which mainly consists of self-attention modules and feed-forward modules. In this transformer encoder, the knowledge interacts with data tokens and INR tokens. The learned INR tokens are mapped to the INR weights through FCs, denoted as FC$^{*}$. These INR weights form our INR continuous function, which is specifically designed for time series data. The designed function takes a batch of timestamps $\{t\}^{T}_{i=1}$ as input, implicitly learns the trends, seasonality, and residual information of the given time series data, and finally reconstructs the input signal. The details of the designed form of INR continuous function and the applied frozen pre-trained LLM encoder module are demonstrated in Section \ref{inr_function} and Section \ref{pretrained_llm}. The anomaly criterion is introduced in Section \ref{anomaly_criterion}.

\vspace{-0.5em}
\subsection{Form of INR Continuous Function}
\label{inr_function}
As shown in Figure \ref{fig_workflow}, we innovatively propose a INR continuous function to better learn and reconstruct time series data. Inspired by classical time series decomposition methods \cite{cleveland1990stl,fons2022hypertime}, the proposed INR continuous function $f$ consists of three components, including trend $f_{tr}$, seasonal $f_{s}$, and residual $f_{r}$:
\begin{equation}
\label{eq_total}
    f(t) = f_{tr}(t) + f_{s}(t) + f_{r}(t).
\end{equation}

\begin{figure}[t]
\centering
\includegraphics[width = 8.4cm]{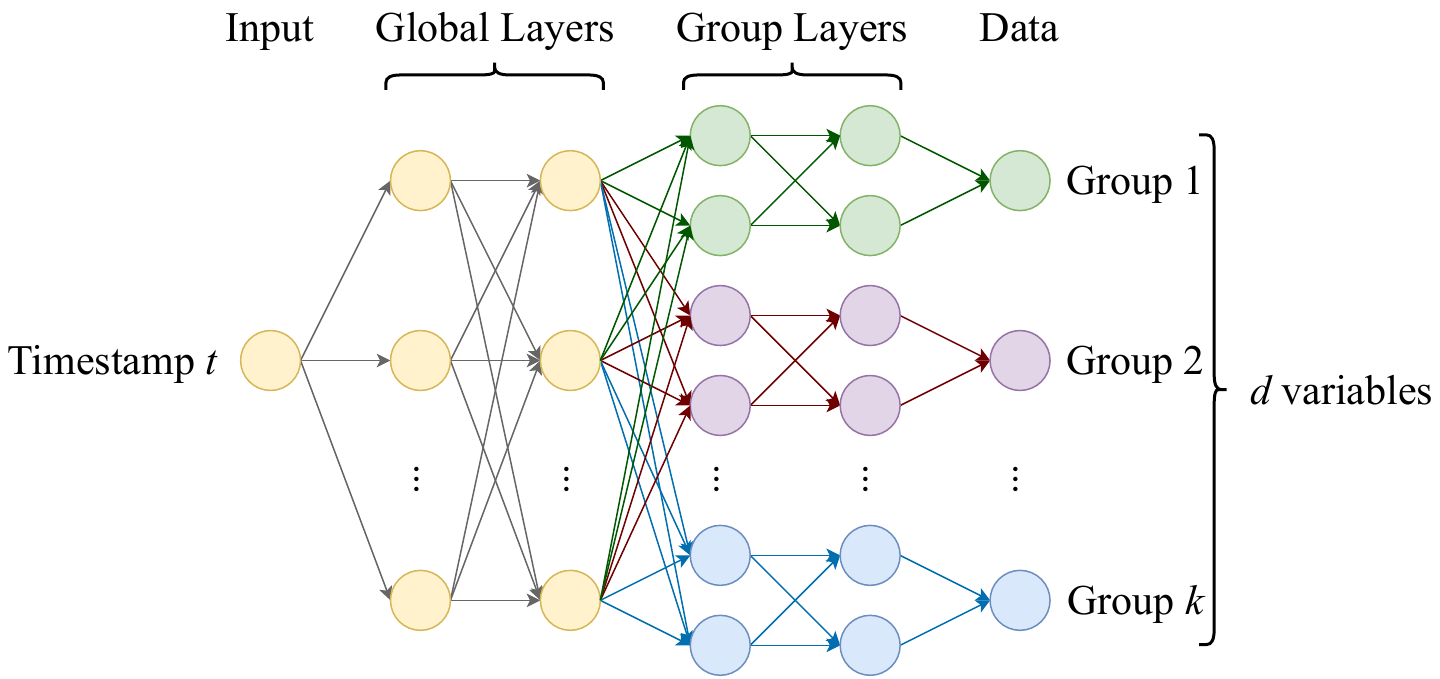}
\caption{A sample of the proposed group-based architecture. It contains 2 global layers and 2 group layers. In this case, each group corresponds to one variable.}
\label{fig_group_mlp}
\vspace{-1em}
\end{figure}

The trend component captures the underlying long-term patterns and focuses on slowly varying behaviors. In order to model this monotonic function, a polynomial predictor is applied \cite{oreshkin2019nbeats,fons2022hypertime}:
\begin{equation}
\label{eq_trend}
    f_{tr}(t)=\sum_{i=0}^p \boldsymbol{w}_{tr}^{(i)} t^{i},
\end{equation}
where $\boldsymbol{w}_{tr}^{(i)}$ is the polynomial coefficients corresponding to the $i^{th}$ degree and $\boldsymbol{w}_{tr}^{(i)}$ is predicted by a FC network. In addition, $p$ denotes the polynomial degree and is set to be small in order to model the low-frequency information and mimic the trend.

The seasonal component grasps the regular, cyclical, and recurring short-term fluctuations. Therefore, a periodic function is employed based on Fourier series \cite{oreshkin2019nbeats,fons2022hypertime}:
\begin{equation}
\label{eq_sea}
    f_{s}(t)=\sum_{i=0}^{\lfloor T / 2-1\rfloor}\left(\boldsymbol{w}^{(i)}_{s} \cos (2 \pi i t)+\boldsymbol{w}^{(i + \lfloor T / 2 \rfloor)}_{s} \sin (2 \pi i t)\right),
\end{equation}
where $\boldsymbol{w}_{s}$ are Fourier coefficients learned by a FC network. This component is then able to model the periodic information and simulate typical seasonal patterns.

The residual component aims to represent the unexplained variability in the data after accounting for the trend and seasonal components. In order to capture this complex and non-periodic information, we design a group-based architecture as shown in Figure \ref{fig_group_mlp}. For any given timestamp $t$, we design $M$ global layers and $N$ group layers with $k$ groups. The global layers capture the inter-channel information while the group layers focus on intra-channel information. The equation for calculation within global layers can be defined as:
\begin{equation}
\label{eq_glo}
q_{m+1}=\mathrm{ReLU} \left(\boldsymbol{w}^{(m)}_{r} q_m+\boldsymbol{b}^{(m)}_{r}\right),
\end{equation}
where $m \in [0,M)$, and $q_m$, $\boldsymbol{w}^{(m)}_{r}$, $\boldsymbol{b}^{(m)}_{r}$ denote the outputs, weights, and biases of the $m^{th}$ global layer respectively. For the group layers, we clone the parameters of $q_m$ for $n$ copies and get $\{q_{M,i}\}_{i=1}^{n}$, each of which is served as the input of the group layers of a group. Then the equation of the output of the $i^{th}$ group at the $(l+1)^{th}$ group layer is given as: 
\begin{equation}
\label{eq_gro}
q_{M+l+1,i}=\mathrm{ReLU} \left(\boldsymbol{w}^{(M+l, i)}_{r} q_{M+l,i}+\boldsymbol{b}^{(M+l,i)}_{r}\right),
\end{equation}
where $l\in[0,N)$, and $q_{M+l}$, $\boldsymbol{w}^{(M+l)}_{r}$, $\boldsymbol{b}^{(M+l)}_{r}$ denote the outputs, weights, and biases of the $l^{th}$ group layer respectively. Finally, the outputs of the $N^{th}$ group layers are concatenated:
\begin{equation}
    f_r(t)= q_{M+N,1} \oplus q_{M+N,2} \oplus \ldots \oplus q_{M+N,n}.
\end{equation}

\subsection{Frozen Pre-trained LLM Encoder}
\label{pretrained_llm}

To further enhance the ability of TSINR to detect anomalies, we leverage the representational capability of LLM. A pre-trained LLM is employed as the encoder, which has been demonstrated to process time-series data and provide cross-modal knowledge \cite{zhou2023one}. With this pre-trained LLM, we map the input data into the feature domain to amplify the fluctuations of anomalies from both time and channel dimensions. On the one hand, in the time dimension, we observe that the extracted feature of LLM involves more intense fluctuations during the anomaly interval. On the other hand, in the channel dimension, other channels have the same anomaly interval due to the ability of LLM to extract and fuse the inter-channel information. Therefore, TSINR can exhibit greater sensitivity to anomalous data, thereby enhancing its ability for anomaly detection. The corresponding experimental results and analysis can be found in Section \ref{abla_llm}.

More specifically, the self-attention layers and the feed-forward layers are frozen to preserve the prior knowledge in the pre-trained model. For any given time series data $X \in \mathbb{R}^{d \times T}$, the pre-trained LLM encoder maps it to feature domain:
\begin{equation}
    Z = Encoder(X),
\end{equation}
where $Z \in \mathbb{R}^{d \times T}$ denotes the feature corresponding to $X$.

\subsection{Anomaly Criterion}
\label{anomaly_criterion}
Following previous reconstruction-based anomaly detection approaches \cite{sakurada2014anomaly,park2018multimodal,zhou2023one}, we use the reconstruction error as the anomaly score for each time series data point. The anomaly score at timestamp $t$ is defined as follows:
\begin{equation}
    AnomalyScore(t) = \frac{1}{d} \sum_{i=1}^d\left\|x_{t,i}-f(t)_i\right\|^2.
\end{equation}
Based on this point-wise anomaly score, we use a parameter threshold $\delta$ to determine whether the point is abnormal or normal:
\begin{equation}
    y_t= \begin{cases}1: \text { abnormal } & \text {AnomalyScore}\left(t\right) \geq \delta \\ 0: \text { normal } & \text {AnomalyScore}\left(t\right)<\delta\end{cases}.
\end{equation}
The threshold $\delta$ is set to label a proportion $\gamma$ of the test dataset as anomalies. And $\gamma$ is a hyper-parameter based on actual datasets.

\begin{table*}[]
\caption{The overall results on multivariate and univariate datasets. The precision (P), recall (R), and F1-score (F1) values are reported, all in percentage (\%). The best results of F1 on each dataset and the average P, R F1 among all datasets are in Bold and the second best ones are \underline{underlined}.} 
\vspace{-0.5em}
\centering
\begin{tabulary}{\linewidth}{c|c|ccccccccccc|c}
\hline
Dataset                  & Metric & \centering Trans. & \centering FED.      & \centering Anomaly. & \centering Auto.     & \centering Pyra. & \centering In. & \centering ETS.      & \centering LightTS        & \centering Dlinear & \centering TimesNet       & \centering FPT   & TSINR          \\ \hline
\centering \multirow{3}{*}{SMD}     & P      & \centering 78.32       & \centering 78.45          & \centering 78.72        & 78.49          & 78.49      & 78.37    & 86.63          & 87.04          & 87.27   & 87.98 & 87.27 & 83.09          \\
                         & R      & 65.24       & 65.08          & 65.43        & 65.13          & 65.53      & 65.23    & 75.35          & 78.39          & 80.99   & 81.54 & 81.08  & 80.46          \\
                         & F1     & 71.19       & 71.14          & 71.46        & 71.19          & 71.43      & 71.20    & 80.68          & 82.49          & 84.01   & \textbf{84.64} & \underline{84.06}  & 81.76          \\ \hline
\centering \multirow{3}{*}{PSM}     & P      & 90.75       & 99.99           & 98.76        & 99.99 & 99.62      & 99.68    & 98.17          & 98.29          & 98.66   & 98.51          & 98.55 & 99.21          \\
                         & R      & 54.68       & 81.89          & 83.25        & 78.99          & 88.46      & 83.30    & 91.36          & 93.60          & 94.70   & 96.27 & 95.79  & 89.37          \\
                         & F1     & 68.24       & 90.04          & 90.35        & 88.26          & 93.71      & 90.75    & 94.64          & 95.89          & 96.64   & \textbf{97.38} & \underline{97.15}  & 94.04          \\ \hline
\multirow{3}{*}{SWaT}    & P      & 99.67       & 99.95           & 99.73        & 99.96 & 99.71      & 99.64    & 92.01          & 92.36          & 92.25   & 92.14          & 92.12 & 99.31          \\
                         & R      & 68.93       & 65.56          & 68.07        & 65.56          & 68.05      & 68.96    & 93.33 & 93.32           & 93.10   & 93.09          & 93.06 & 72.32          \\
                         & F1     & 81.50       & 79.19          & 80.91        & 79.19          & 80.90      & 81.51    & 92.67          & \textbf{92.84} & \underline{92.68}    & 92.61          & 92.59 & 83.69          \\ \hline
\multirow{3}{*}{MSL}     & P      & 90.58       & 90.69 & 89.78        & 90.66           & 90.64      & 90.63    & 86.89          & 89.17          & 89.68   & 89.55          & 82.03 & 83.57          \\
                         & R      & 74.65       & 75.48          & 73.66        & 75.22          & 74.76      & 74.96    & 67.78          & 73.64          & 75.31   & 75.29          & 82.01  & 85.40 \\
                         & F1     & 81.85       & \underline{82.39}           & 80.93        & 82.22          & 81.94      & 82.06    & 76.16          & 80.66          & 81.87   & 81.80          & 82.02 & \textbf{84.47} \\ \hline
\multirow{3}{*}{SMAP}    & P      & 90.87       & 89.98          & 90.14        & 90.72          & 89.51      & 90.66    & 90.75          & 90.02          & 89.89   & 89.92          & 90.91  & 91.67 \\
                         & R      & 61.44       & 55.89          & 54.00        & 62.58           & 54.59      & 61.69    & 54.68          & 53.90          & 54.01   & 56.56          & 61.01 & 76.42 \\
                         & F1     & 73.31       & 68.95          & 67.54        & \underline{74.07}           & 67.82      & 73.43    & 68.24          & 67.43          & 67.48   & 69.44          & 73.02 & \textbf{83.35} \\ \hline
\multirow{3}{*}{PTB-XL}    & P      & 56.89       & 48.36          & 56.00        & 49.12          & 50.22      & 57.41    & 62.84          & 66.38          & 62.95   & 67.60          &  71.85  & 58.35 \\
                         & R      & 29.99       & 27.60          & 31.50        & 27.53           & 23.85      & 25.43    & 28.45          & 16.46          & 13.95   & 14.47          & 24.52 & 35.00 \\
                         & F1     & 39.28       & 35.14          & \underline{40.32}        & 35.28           & 32.34      & 35.25    & 39.17          & 26.38          & 22.84   & 23.84          & 36.57 & \textbf{43.75} \\ \hline
\multirow{3}{*}{SKAB}     & P      & 87.56       & 86.88          & 91.83         & 87.51          & 89.55      & 88.67    & 85.38          & 83.83          & 86.01   & 85.65          & 86.18 & 89.98 \\
                         & R      & 86.72       & 77.71          & 95.04        & 91.10          & 97.27      & 97.27     & 100.00          & 82.01          & 100.00   & 100.00          & 99.21 & 98.65 \\
                         & F1     & 87.14       & 82.04          & \underline{93.41}         & 89.27          & 93.25      & 92.77    & 92.12          & 82.91          & 92.48   & 92.27          & 92.24 & \textbf{94.11} \\ \hline
\multirow{3}{*}{UCR}     & P      & 41.13       & 32.96          & 44.79         & 42.82          & 42.12      & 43.97    & 40.12          & 37.70          & 34.55   & 33.11          & 41.00 & 67.29 \\
                         & R      & 33.61       & 25.73          & 34.83        & 33.97          & 35.13      & 35.16     & 29.85          & 29.01          & 29.06   & 29.18          & 32.51 & 62.35 \\
                         & F1     & 34.50       & 27.09          & \underline{36.51}         & 35.52          & 36.02      & 36.41    & 31.94          & 30.82          & 29.67   & 29.81          & 34.33 & \textbf{62.46} \\ \hline
\multirow{3}{*}{Average} & P      & 79.47       & 78.41          & 82.22        & 79.91          & 79.98      & 81.13     & 80.35          & 80.60          & 80.16   & 80.56          & \underline{81.24} & \textbf{84.06} \\
                         & R      & 59.41       & 59.37          & 63.22        & 62.51          & 63.46      & 64.00    & 67.60          & 65.04          & 67.64   & 68.30          & \underline{71.15}  & \textbf{75.00} \\
                         & F1     & 67.13       & 67.00          & 70.18        & 69.38          & 69.68      & 70.42    & 71.95          & 69.93          & 70.96   & 71.47          & \underline{74.00}  & \textbf{78.45} \\
                         \hline
\end{tabulary}
\label{main_results}
\vspace{-0.2em}
\end{table*}

\section{Experiments}
\subsection{Datasets}
We use eight anomaly detection benchmarks from real-world scenarios to validate the performance of our proposed method, including seven multivariate datasets (\textbf{SMD} \cite{su2019robust}, \textbf{PSM} \cite{abdulaal2021practicalPSM}, \textbf{SWaT} \cite{mathur2016swat}, \textbf{MSL} \cite{hundman2018detecting}, \textbf{SMAP} \cite{hundman2018detecting}, \textbf{PTB-XL} \cite{jiang2023multi-ecgad, wagner2020ptbxl}, and \textbf{SKAB} \cite{skab}) and one univariate dataset (\textbf{UCR} \cite{wu2021currentUCR}).

\subsection{Baselines and Experimental Settings}
We compare our proposed method with 11 state-of-the-art deep learning approaches, including both general frameworks designed for time series modeling and algorithms specifically tailored for time series anomaly detection: FPT \cite{zhou2023one}, TimesNet \cite{wu2022timesnet}, ETSformer \cite{woo2022etsformer}, FEDformer \cite{zhou2022fedformer}, LightTS \cite{zhang2022less}, DLinear \cite{zeng2023transformers}, Autoformer \cite{wu2021autoformer}, Pyraformer \cite{liu2021pyraformer}, AnomalyTransformer \cite{xu2021anomalytransformer}, Informer \cite{zhou2021informer} and Transformer \cite{vaswani2017attention}. The commonly used metrics of precision (P), recall (R), and F1-score are employed for evaluation. Additionally, we report threshold-free measurements, including the Area Under the Curve (AUC) and Volume Under the Surface (VUS) \cite{paparrizos2022volume}, which are provided in the Appendix \ref{app_addition}.

The implementation details and the default hyper-parameters are summarized here. For a fair comparison, we only employ the classical reconstruction error across all baseline models. Also, we adopt identical data processing methods and the corresponding parameter configurations. We employ the sliding window approach and use a fixed window size of 100 for all datasets. The proportion $\gamma$ mentioned in Section \ref{anomaly_criterion} is set to 0.5 for SMD dataset, 0.1 for UCR dataset, 10 for SKAB dataset, and 1 for others. These parameters adhere to the settings of previous work \cite{zhou2023one,xu2021anomalytransformer}. Ablation studies on the anomaly proportion $\gamma$ are in Appendix \ref{app_anomalyp}. For the main results, our TSINR model involves 3 global layers and 2 group layers in the residual block. And the hidden dimensions are 64 and 32 respectively. The transformer encoder has 6 blocks. We use GPT2 \cite{radford2019language} as the pre-trained LLM encoder and 6 blocks are utilized following the same settings as in FPT \cite{zhou2023one}. The experiments are conducted using the ADAM optimizer \cite{kingma2014adam} with an initial learning rate of $10^{-4}$. A single NVIDIA Tesla-V100 32GB GPU is applied for each dataset. And the efficiency analysis is in Appendix \ref{app_efficiency}.

\vspace{-0.5em}
\subsection{Main Resutls}
We compare our method with 11 other state-of-the-art approaches and the results are shown in Table \ref{main_results}. These results show that our method achieves superior overall performance on these benchmark datasets. These experimental results confirm that TSINR, in both multivariate and univariate scenarios, effectively captures temporal continuity and precisely identifies discontinuous anomalies. The findings affirm the robustness of TSINR across diverse datasets and showcase its potential for broader applications in diverse domains.

In multivariate scenarios, we observe that despite both MSL and SMAP being collected from NASA Space Sensors, TSINR achieved significantly greater improvements on the SMAP dataset compared to other methods. This could be attributed to the presence of more point anomalies in the SMAP dataset. Point anomalies exhibit poorer continuity compared to other anomaly patterns due to their isolated nature, representing single data points that significantly deviate from the surrounding pattern. This aligns with the property of spectral bias, making our model more sensitive to point anomalies, thereby achieving greater improvements on the SMAP dataset. Meanwhile, our approach shows moderate performance on the SWAT dataset. This is due to the poorer continuity of the SWaT dataset, which affects the fitting ability of TSINR.

The situation is more complex in the UCR dataset. The UCR dataset comprises 250 univariate sub-datasets from various domains and we report the average scores to provide a comprehensive assessment. The results demonstrate that our approach still outperforms other methods in overall performance by a significant margin. These sub-datasets originate from various domains, demonstrating the generalization capability of our method. In addition, it proves the strong ability of TSINR for modeling and identifying anomalies in univariate scenarios and underscores the importance of the spectral bias constraint for anomaly detection.

\begin{table*}[]
\centering
\caption{Ablation studies on the decomposition components and the group-based architecture. The F1 score is reported and the best results are in \textbf{Bold}.}
\vspace{-0.3em}
\begin{tabular}{cccccccccc}
\hline
Decomposition & Group-based & SMD   & PSM   & SWaT  & MSL   & SMAP & PTB-XL & SKAB  \\ \hline
\ding{55}            & \ding{55}          & 78.52 & 92.61 & 81.89 & 82.02 & 73.31 & 40.11 & 93.13 \\
\ding{55}            & \ding{51}          & 80.10 & 93.08 & 82.28 & 82.28 & 78.94 & 40.76 & 94.01 \\
\ding{51}             & \ding{55}           & 79.24 & 93.04 & 82.16 & 82.95 & 78.66 & 42.65 & 93.96 \\
\ding{51}             & \ding{51}           & \textbf{81.76} & \textbf{94.04} & \textbf{83.69} & \textbf{84.74} & \textbf{83.35} & \textbf{43.75} & \textbf{94.11} \\ \hline
\end{tabular}
\vspace{-0.3em}
\label{ablation_group}
\end{table*}

\begin{table*}[]
\centering
\caption{Analysis of the decomposition components. The MSE and F1 scores (*/*) are reported and the best results are in \textbf{Bold}.}
\vspace{-0.3em}
\begin{tabular}{ccccccc}
\hline
Decomposition & Synthetic Trend    & Synthetic Seasonal & SMD                 & PSM                 & MSL                 & SKAB                \\ \hline
\ding{55}             & 0.56/71.15        & 0.44/25.97        & 1.21/80.10          & 0.26/93.08          & 2.23/82.28          & 1.23/94.01          \\
\ding{51}             & \textbf{0.09/100.00} & \textbf{0.01/100.00} & \textbf{0.99/81.76} & \textbf{0.21/94.04} & \textbf{1.81/84.74} & \textbf{0.84/94.11} \\ \hline
\end{tabular}
\vspace{-0.3em}
\label{ablation_decom}
\end{table*}

\begin{table*}[]
\centering
\caption{Ablation studies on the pre-trained LLM encoder. The F1 score is reported and the best results are in \textbf{Bold}.}
\vspace{-0.3em}
\begin{tabular}{ccccccccc}
\hline
Pre-trained LLM & SMD   & PSM   & SWaT  & MSL   & SMAP & PTB-XL & SKAB & UCR  \\ \hline
\ding{55}       & 80.29 & 92.69 & 82.33 & 83.27 & 79.35 & 40.04 & 93.91 & \textbf{62.46} \\
\ding{51}       & \textbf{81.76} & \textbf{94.04} & \textbf{83.69} & \textbf{84.47} & \textbf{83.35} & \textbf{43.75} & \textbf{94.11} & 60.41 \\
\hline
\end{tabular}
\label{ablation_llm}
\end{table*}

\vspace{-0.8em}
\subsection{Ablation Studies}

\subsubsection{Analysis of the Decomposition Components and the Group-based Architecture}

In this section, we analyze the effectiveness of the proposed decomposition components and group-based architecture. The decomposition components indicate the three components (i.e., trend, seasonal, and residual) designed in our paper. And the group-based architecture is proposed for the residual block. 

\begin{figure}[]
\centering
\includegraphics[width = \linewidth]{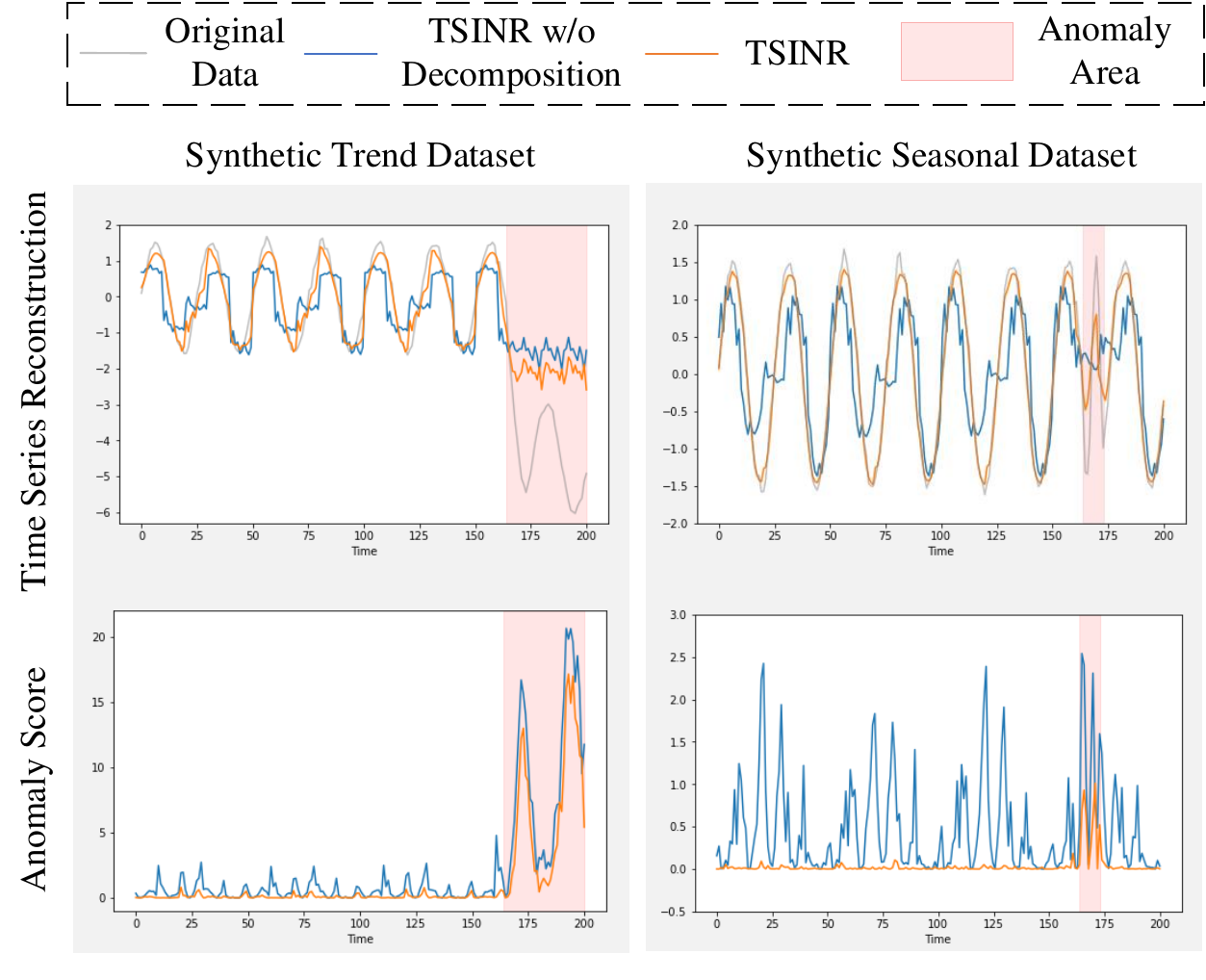}
\caption{The visual analysis of the decomposition components on synthetic trend and seasonal datasets.}
\label{fig_trend}
\vspace{-0.5em}
\end{figure}

\begin{figure}[]
\centering
\includegraphics[width = \linewidth]{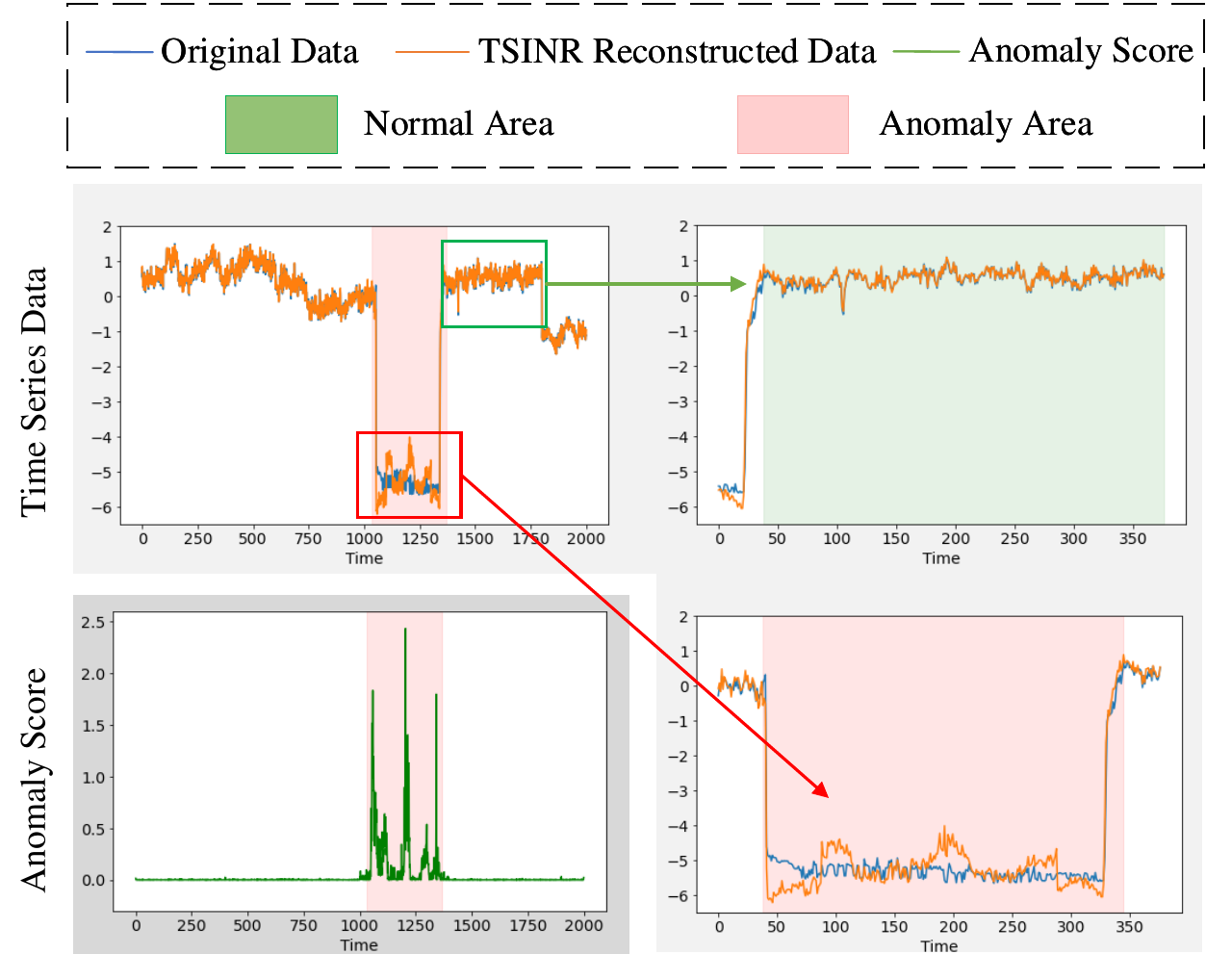}
\caption{The visual analysis of the non-spike anomaly segment in the real-world PSM dataset.}
\label{fig_nonspike}
\vspace{-0.5em}
\end{figure}

The main purpose of the decomposition components is to extract the unique trend and seasonal information of the time series data. The results in Table \ref{ablation_group} indicate that capturing these distinctive features significantly enhances the capability for anomaly detection. To further explain the effectiveness of the trend and seasonal components in INR continuous function, we conduct ablation experiments using both synthetic and real-world datasets. The synthetic trend and seasonal datasets are generated by \cite{lai2021revisiting}, while the real-world datasets include SMD, PSM, MSL, and SKAB. As shown in Table \ref{ablation_decom} and Figure \ref{fig_trend}, incorporating trend and seasonal components significantly improves both reconstruction performance and anomaly detection capability. In contrast, omitting these components results in inadequate data fitting in certain cases, thereby hindering the ability to detect anomalies. Also, we showcase the detection of a non-spike anomaly segment in Figure \ref{fig_nonspike}, highlighting the role of trend and seasonal components in identifying such anomalies. The anomaly, characterized by subtle deviations from the expected behavior, is effectively captured, demonstrating the ability of TSINR to detect anomalies that do not exhibit abrupt or spike-like changes. This emphasizes the robustness of the model in handling different types of anomalies in real-world datasets like PSM.

\begin{figure*}[]
\centering
\includegraphics[width = \textwidth]{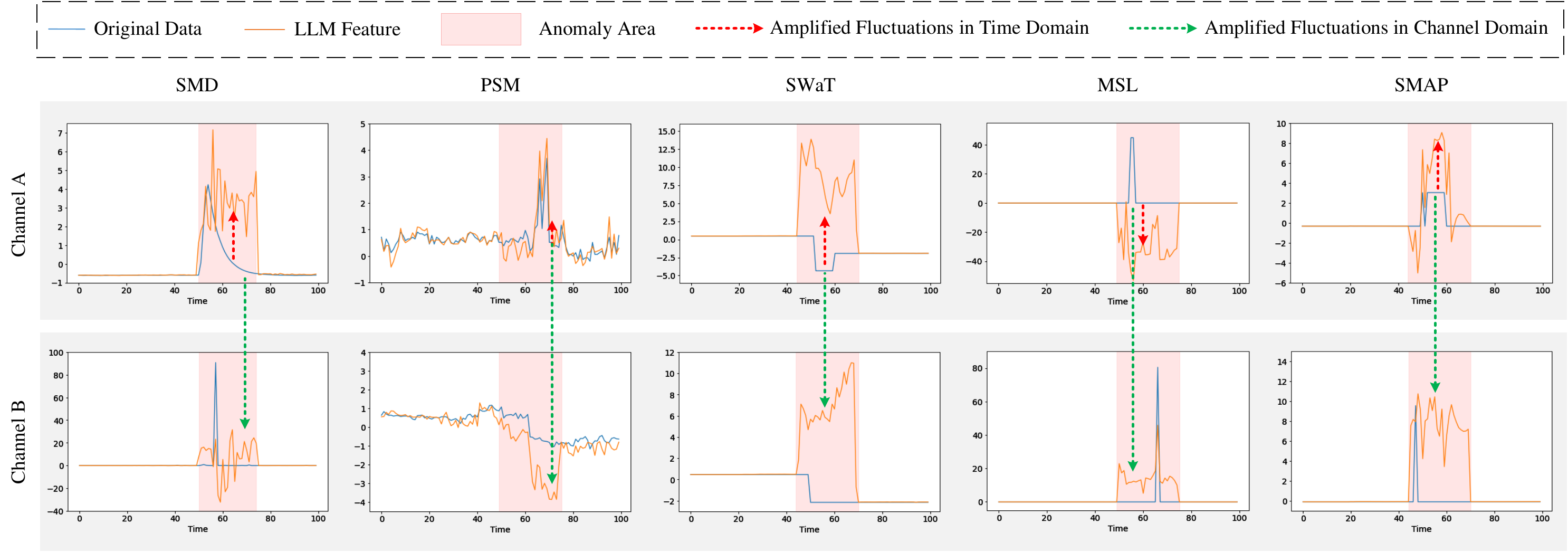}
\vspace{-1.5em}
\caption{The visualization of the original data and the corresponding features obtained from the frozen pre-trained LLM encoder. The intense fluctuations in the anomaly area are amplified from both time and channel domains.}
\label{fig_LLM_feature}
\vspace{-1.2em}
\end{figure*}

\begin{figure}[]
\centering
\subfloat[]
{
 \centering
 \includegraphics[width=4cm]{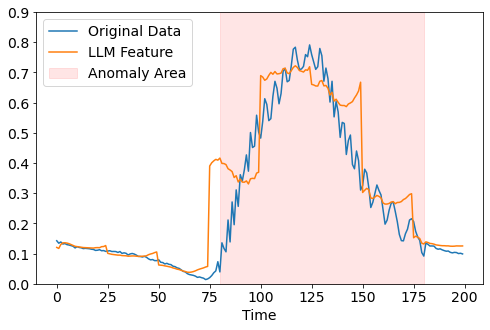}
 \label{fig_LLM_feature1}
}
\hfill
\subfloat[]
{
 \centering
 \includegraphics[width=4cm]{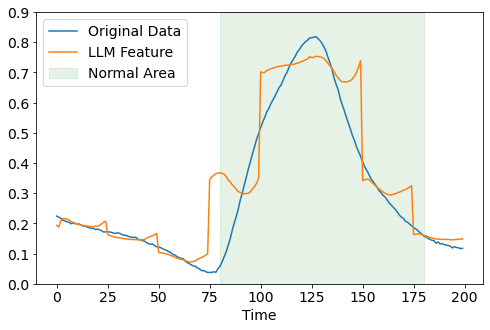}
 \label{fig_LLM_feature2}
}
\vspace{-0.8em}
\caption{The visualization of the UCR original data and the corresponding features obtained from the frozen pre-trained LLM encoder of (a) anomaly area and (b) normal area.} 
\label{fig_LLM}
\vspace{-2em}
\end{figure}

In addition, the group-based architecture is designed to enhance the representational capacity of INR for multivariate data. Experimental results indicate an improvement in the capability for anomaly detection when employing the proposed group-based architecture. This is because modeling multiple variables and capturing both inter- and intra-channel information with a simple continuous function, which only consists of fully-connected layers, is challenging. Our approach addresses this by dividing the variables into several groups and applying independently fully-connected layers in different groups, thereby reducing the number of variables each group needs to model and improving the representational capacity. The global layers extract the inter-channel information, while the group layers selectively focus on detailed information for specific channels. This enhances the representational capability for each variable without losing any knowledge. Detailed ablation studies on the number of groups are left in Appendix \ref{app_group_num}.

\vspace{-0.5em}
\subsubsection{Analysis of Pre-trained LLM Encoder}
\label{abla_llm}

Further, we prove the validity of the pre-trained LLM encoder, which is utilized to encode the data into the feature domain to amplify the fluctuations of anomalies and thereby enhance the capability of TSINR in identifying anomalies. Table \ref{ablation_llm} displays the ablation studies of the pre-trained LLM encoder. For the multivariate datasets, it can be observed that applying this encoder enhances the performance of anomaly detection. 
To further demonstrate the effectiveness, we compare the raw data with the features extracted through the encoder. As shown in Figure \ref{fig_LLM_feature}, the figures in the first row illustrate that during the time interval when anomalies occur, the extracted features exhibit more pronounced fluctuations compared to the original data. This implies that the discontinuity in anomalies is increased in time domain. Also, these extracted features incorporate inter-channel information, providing a manifestation of anomalies among all variables. As shown in the second line, the features exhibit anomalous fluctuations in the same time interval as other channels, whereas the original data only shows a brief peak. This verifies that the anomalies are shared in channel domain. Based on these results, we indicate that utilizing the pre-trained LLM encoder can effectively enhance abnormal information both intra- and inter-channel. This aligns with the spectral bias of INR, making our model more sensitive to anomalous data.

In contrast, on the UCR dataset, using the pre-trained LLM encoder actually decreased the performance of anomaly detection. This is because the advantages observed in the aforementioned multivariate datasets do not apply to the UCR dataset. On the one hand,  there are no anomalies in the UCR training data. Consequently, during inference, the LLM encoder still extracts features according to the normal pattern. As shown in Figure \ref{fig_LLM}, the highly discontinuous anomalous data becomes relatively smooth after feature extraction. Meanwhile, there is little variation in features between normal and abnormal areas. This prompts TSINR to fit the anomalous data, thereby reducing its sensitivity to anomalies. On the other hand, the UCR dataset involves only a single variable, thus the inter-channel information provided by the LLM encoder is meaningless.

\begin{figure*}[ht]
\centering
\includegraphics[width = \textwidth]{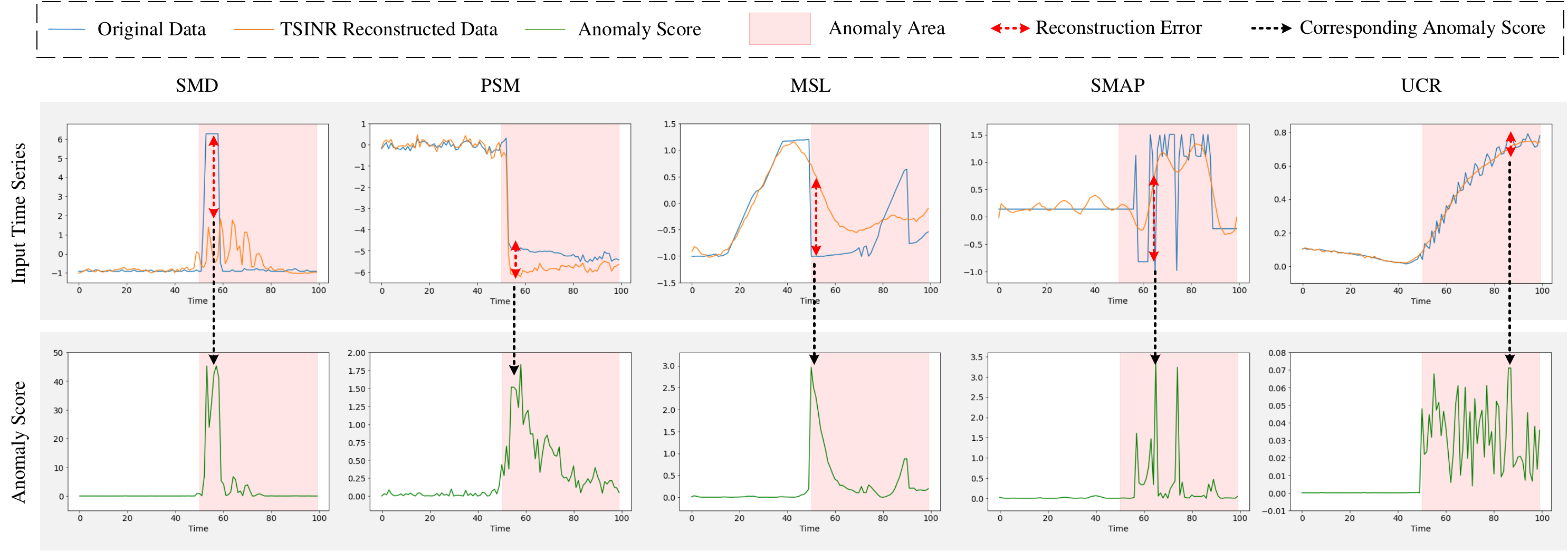}
\caption{The visualization of the original data, the TSINR reconstructed data, and the corresponding anomaly score. In the anomaly area, anomaly scores significantly increase, demonstrating the sensitivity of TSINR to anomalies.}
\label{fig_TSINR_feature}
\vspace{-0.5em}
\end{figure*}

\begin{figure*}[]
\centering
\includegraphics[width = \textwidth]{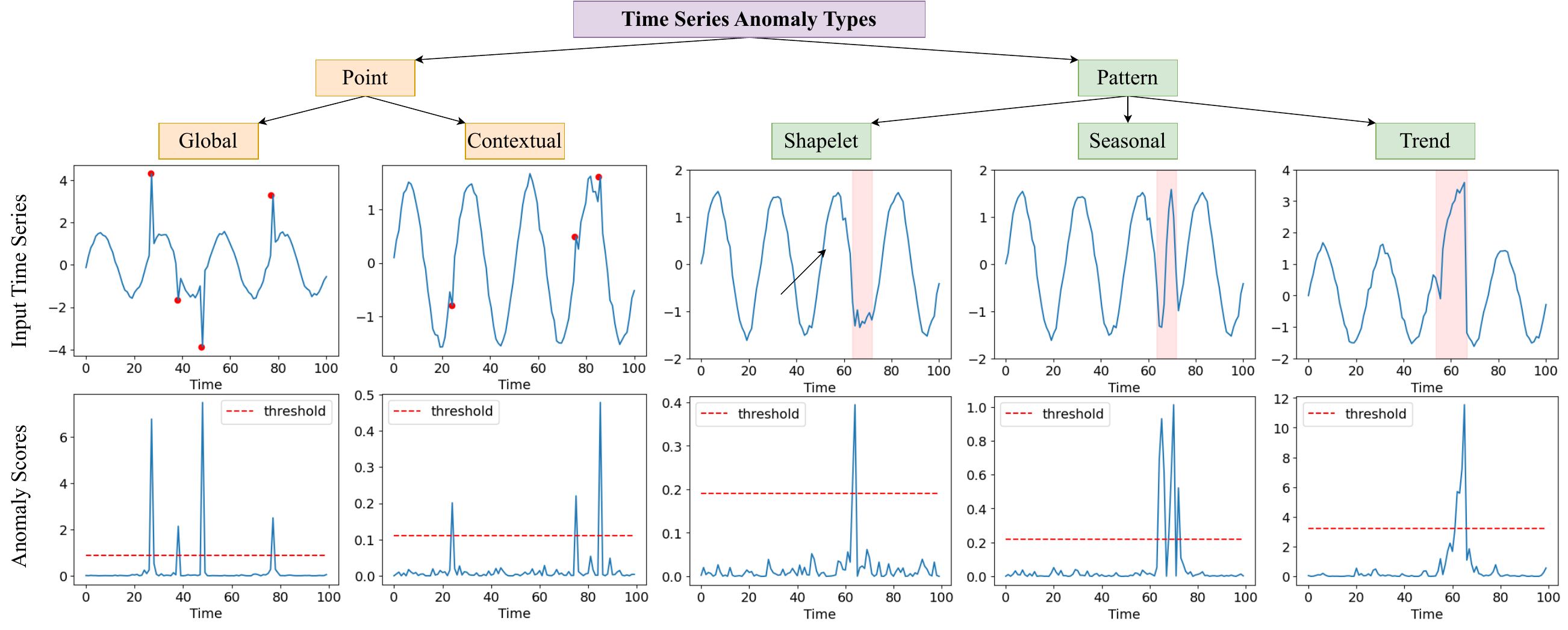}
\vspace{-2em}
\caption{The visualization of the ground-truth anomalies and anomaly scores of TSINR for different types of anomalies.}
\label{fig_type_anomalies}
\vspace{-1em}
\end{figure*}

\subsection{Visual Analysis}
In order to demonstrate that our approach is sensitive to discontinuous abnormal data, 
we compare the original data with the reconstructed values of the TSINR model. As shown in Figure \ref{fig_TSINR_feature}, the smooth normal points are well-fitted, while the discontinuous abnormal points are not. The anomaly scores significantly increase when anomalies occur, which aids the TSINR model in distinguishing abnormal points, demonstrating its sensitivity to the anomaly points. It is worth noting that, despite the small and highly fluctuating values of anomaly scores on the UCR dataset, this still holds meaningful significance. The UCR dataset involves only one anomaly, hence even minor fluctuations are helpful for the model to pinpoint these anomaly points.

Furthermore, we validate the robustness of TSINR with the synthetic data generated for time series anomaly detection. It has univariate time series and involves different types of anomalies \cite{yang2023dcdetector,lai2021revisiting}, including the point-wise anomaly (global point and contextual point anomalies) and pattern-wise anomalies (shapelet, seasonal, and trend anomalies). As shown in Figure \ref{fig_type_anomalies}, the
red points are anomaly points and the red areas are anomaly areas. It can be seen that TSINR can robustly detect various types of anomalies from normal points with relatively high anomaly scores.

\section{Conclusion}
Time series data anomaly detection plays a pivotal role in ensuring the reliability, security, and efficiency of systems across various domains. Reconstruction-based methods are mainstream approaches for this task because they do not require label information and are easy to interpret the detection results. However, unlabeled anomalous points in the training data can negatively impact the performance of these reconstruction models. To address this issue, this paper proposes a novel algorithm named TSINR for time series anomaly detection based on INR reconstruction. We utilize the spectral bias of INR to prioritize fitting continuous normal data and capture temporal continuity, thus enhancing sensitivity to discontinuous anomalous data. A transformer-based architecture is employed to predict the parameters of INR. To cope with the complex patterns of time series data, we specifically design a formulation of continuous function. It aims to implicitly learn the trend, seasonal, and residual information and capture the inter- and intra-channel information of the time series data. Besides, we leverage a frozen pre-trained LLM encoder to map the original data to the feature domain, thus amplifying the fluctuations of anomalies from both time and channel domains and enabling TSINR to better distinguish between abnormal and normal points. Experimental results indicate that TSINR exhibits superior overall performance on both multivariate and univariate benchmark datasets compared to other state-of-the-art algorithms. Also, ablation studies verify the effectiveness of each component, and visual analysis demonstrates the sensitivity of TSINR to anomalous points. In this work, we demonstrate the potential of INR in time series data tasks. In future work, we plan to explore the performance of INR on other time series tasks, including imputation, classification, and long-term and short-term forecasting. We believe that INR has the potential to become a unified framework for various time-series data tasks.

\section{Acknowledgments}

This work was supported by the National Natural Science Foundation of China (Grant No. 62202422), the National Key R\&D Program of China (Grant No. 2022ZD0160703), the National Natural Science Foundation of China (Grant Nos. 62276230 and 62372408), Zhejiang Provincial Natural Science Foundation of China (Grant No. LDT23F02023F02), and Shanghai Artificial Intelligence Laboratory.

\newpage
\bibliographystyle{ACM-Reference-Format}
\bibliography{sample-base}

\appendix

\lstset{
    language=Python, 
    basicstyle=\ttfamily\small, 
    keywordstyle=\color{blue}, 
    stringstyle=\color{red}, 
    commentstyle=\color{green!70!black}, 
    showstringspaces=false, 
    breaklines=true, 
    numbers=left, 
    numberstyle=\tiny\color{gray}, 
    captionpos=b 
}
\section*{Appendix}
\section{Additional Results}
\label{app_addition}
In addition to the commonly used F1-score, which is threshold-dependent, we evaluate TSINR using threshold-free AUC and VUS scores. These metrics provide a more comprehensive assessment of model performance, as they do not rely on a specific decision threshold. The results in Table demonstrate that TSINR consistently outperforms competing methods across all datasets, highlighting its robustness and superior capability in anomaly detection tasks.

\section{Study on the Anomaly Proportion}
\label{app_anomalyp}
Anomaly proportion $\gamma$ is a hyper-parameter which decides the anomaly threshold $\delta$. As mentioned in Section \ref{anomaly_criterion}, the threshold $\delta$ is set to label a proportion $\gamma$ of the test dataset as anomalies. We show the influence of the anomaly proportion in Table \ref{app_table_anomalyp}. It can be observed that an appropriate anomaly proportion is beneficial in aiding the model's judgment of anomalies. Among them, PSM exhibits greater robustness to anomaly proportion compared to SMD and MSL. This is consistent with previous findings \cite{yang2023dcdetector}.

\section{Efficiency Analysis}
We measure the efficiency of the TSINR method, and show the results in Table \ref{app_tab_effi}. The results indicate that the TSINR is pretty efficient and lightweight.
\label{app_efficiency}

\section{Study on the Group Number}
\label{app_group_num}
The group number is a hyperparameter in our method, indicating the number of groups in the group-based architecture. It determines the number of variables learned by neurons in each group layer. Table \ref{app_table_groupnum} presents ablation studies on the group number. When set to 1, the method reverts to a regular function without grouping. Our results show that the optimal group number varies across datasets, which is reasonable given that, similar to image data, partitioning accelerates INR fitting and is dependent on the dataset’s intrinsic characteristics. Nonetheless, the group-based architecture improves the model's anomaly detection performance.

\begin{table*}[]
\caption{The additional results on multivariate and univariate datasets. The AUC and VUS values are reported. The best results are in Bold and the second best ones are \underline{underlined}.} 
\vspace{-0.5em}
\centering
\begin{tabulary}{\linewidth}{c|c|ccccccccccc|c}
\hline
Dataset                 & Metric & Trans.       & FED.   & Anomaly.     & Auto.  & Pyra.        & In.          & ETS.   & LightTS      & DLinear & TimesNet     & FPT          & TSINR           \\ \hline
\multirow{2}{*}{SMD}    & AUC    & 0.747       & 0.654 & 0.765       & 0.652 & 0.728       & 0.729       & 0.760 & 0.738       & 0.732  & \underline{0.766} & 0.723       & \textbf{0.774} \\
                        & VUS    & 0.740       & 0.639 & 0.762       & 0.631 & 0.723       & 0.722       & 0.755 & 0.734       & 0.728  & \underline{0.762} & 0.720       & \textbf{0.769} \\ \hline
\multirow{2}{*}{PSM}    & AUC    & \underline{0.721} & 0.662 & 0.666       & 0.661 & 0.704       & 0.712       & 0.622 & 0.585       & 0.565  & 0.590       & 0.578       & \textbf{0.722} \\
                        & VUS    & 0.667       & 0.563 & 0.608       & 0.556 & 0.657       & \underline{0.670} & 0.609 & 0.570       & 0.543  & 0.575       & 0.568       & \textbf{0.674} \\ \hline
\multirow{2}{*}{SWaT}   & AUC    & 0.816       & 0.817 & \underline{0.820} & 0.817 & 0.818       & 0.816       & 0.444 & 0.737       & 0.622  & 0.247       & 0.236       & \textbf{0.823} \\
                        & VUS    & 0.518       & 0.512 & 0.530       & 0.521 & 0.534       & 0.515       & 0.435 & \underline{0.701} & 0.598  & 0.241       & 0.232       & \textbf{0.754} \\ \hline
\multirow{2}{*}{MSL}    & AUC    & \underline{0.624} & 0.550 & 0.532       & 0.550 & 0.602       & 0.613       & 0.596 & 0.601       & 0.615  & 0.623       & 0.590       & \textbf{0.657} \\
                        & VUS    & \underline{0.607} & 0.525 & 0.518       & 0.525 & 0.569       & 0.599       & 0.555 & 0.569       & 0.580  & 0.591       & 0.552       & \textbf{0.638} \\ \hline
\multirow{2}{*}{SMAP}   & AUC    & \underline{0.526} & 0.450 & 0.456       & 0.450 & 0.452       & 0.490       & 0.401 & 0.380       & 0.397  & 0.455       & 0.474       & \textbf{0.576} \\
                        & VUS    & \underline{0.502} & 0.418 & 0.450       & 0.415 & 0.438       & 0.482       & 0.363 & 0.343       & 0.373  & 0.412       & 0.444       & \textbf{0.567} \\ \hline
\multirow{2}{*}{PTB-XL} & AUC    & 0.604       & 0.485 & 0.583       & 0.485 & 0.536       & 0.560       & 0.589 & 0.545       & 0.516  & 0.618       & \underline{0.627} & \textbf{0.660} \\
                        & VUS    & 0.456       & 0.339 & 0.436       & 0.339 & 0.386       & 0.417       & 0.453 & 0.401       & 0.365  & 0.471       & \underline{0.486} & \textbf{0.534} \\ \hline
\multirow{2}{*}{SKAB}   & AUC    & 0.536       & 0.429 & 0.493       & 0.440 & \underline{0.570} & 0.496       & 0.482 & 0.480       & 0.504  & 0.496       & 0.496       & \textbf{0.585} \\
                        & VUS    & 0.535       & 0.421 & 0.492       & 0.434 & \underline{0.569} & 0.495       & 0.482 & 0.474       & 0.504  & 0.496       & 0.496       & \textbf{0.585} \\ \hline
\multirow{2}{*}{UCR}    & AUC    & 0.520       & 0.541 & \underline{0.548} & 0.533 & 0.546       & 0.517       & 0.502 & 0.523       & 0.546  & 0.533       & 0.529       & \textbf{0.663} \\
                        & VUS    & 0.511       & 0.513 & \underline{0.538} & 0.508 & 0.533       & 0.506       & 0.486 & 0.511       & 0.528  & 0.514       & 0.513       & \textbf{0.650} \\ \hline
\end{tabulary}
\label{app_tab_auc}
\end{table*}

\begin{table*}[]
\caption{Ablation Studies on the anomaly proportion $\gamma$ which decides the threshold $\delta$. The best results are in Bold.}
\begin{tabular}{c|ccc|ccc|ccc}
\hline
Dataset     & \multicolumn{3}{c|}{SMD}       & \multicolumn{3}{c|}{PSM}       & \multicolumn{3}{c}{MSL}        \\ \hline
Metric      & P     & R     & F1             & P     & R     & F1             & P     & R     & F1             \\ \hline
$\gamma$ = 0.5 & 83.09 & 80.46 & \textbf{81.76} & 99.53 & 88.65 & 93.78          & 94.67 & 61.92 & 74.87          \\
$\gamma$ = 0.6 & 80.55 & 80.93 & 80.74          & 99.37 & 88.90 & 93.85          & 93.16 & 62.36 & 74.71          \\
$\gamma$ = 0.7 & 77.45 & 81.46 & 79.40          & 99.33 & 89.00 & 93.89          & 92.32 & 66.77 & 77.49          \\
$\gamma$ = 0.8 & 75.16 & 83.06 & 78.91          & 99.27 & 88.80 & 93.75          & 91.92 & 73.43 & 81.64          \\
$\gamma$ = 0.9 & 73.44 & 83.37 & 78.09          & 98.82 & 89.03 & 93.67          & 90.46 & 75.20 & 82.12          \\
$\gamma$ = 1.0 & 72.55 & 83.39 & 77.59          & 99.21 & 89.37 & \textbf{94.04} & 83.98 & 84.26 & \textbf{84.12} \\ \hline
\end{tabular}
\label{app_table_anomalyp}
\end{table*}

\begin{table*}[]
\caption{The efficiency of TSINR on the data with 128 batch size.}
\begin{tabular}{c|c|c}
\hline
Training Time per Batch & Inference Time per Batch & Learnable Parameters \\ \hline
0.1s                    & 0.08s                    & 8.3M                 \\ \hline
\end{tabular}
\label{app_tab_effi}
\end{table*}

\begin{table*}[]
\caption{Ablation Studies on the group number of the group-based architecture. The best results are in Bold.}
\begin{tabular}{c|ccc|ccc|ccc}
\hline
Dataset        & \multicolumn{3}{c|}{SMD}                     & \multicolumn{3}{c|}{PSM}                     & \multicolumn{3}{c}{MSL}                      \\ \hline
Metric         & P     & R     & F1                           & P     & R     & F1                           & P     & R     & F1                           \\
\hline
Group Num = 1  & 82.82 & 75.97 & 79.24                        & 99.11 & 87.66 & 93.04                        & 83.30 & 82.59 & 82.95                        \\
Group Num = 2  & 82.22 & 76.61 & 79.31                        & 99.28 & 88.90 & 93.81                        & 83.47 & 82.57 & 83.01                        \\
Group Num = 3  & 82.46 & 77.45 & 79.88                        & 99.25 & 89.02 & 93.86                        & 83.43 & 82.51 & 82.97                        \\
Group Num = 4  & 82.62 & 77.76 & 80.12                        & 98.81 & 89.32 & 93.83                        & 83.60 & 82.76 & 83.17                        \\
Group Num = 5  & 82.29 & 77.30 & 79.72                        & 99.21 & 89.37 & \textbf{94.04} & 83.53 & 82.66 & 83.09                        \\
Group Num = 6  & 83.09 & 80.46 & \textbf{81.76}                        & 99.19 & 88.53 & 93.56                        & 83.57 & 82.71 & 83.14                        \\
Group Num = 7  & 82.71 & 77.05 & 79.78                        & 99.23 & 88.79 & 93.72                        & 83.56 & 82.80 & 83.18                        \\
Group Num = 8  & 82.39 & 77.08 & 79.65                        & 99.41 & 88.57 & 93.68                        & 83.52 & 82.48 & 83.00                        \\
Group Num = 9  & 83.17 & 77.63 & 80.31 & 99.26 & 89.07 & 93.89                        & 83.57 & 85.40 & \textbf{84.47} \\
Group Num = 10 & 82.17 & 76.73 & 79.36                        & 99.30 & 88.81 & 93.77                        & 83.56 & 82.47 & 83.01                        \\ \hline
\end{tabular}
\label{app_table_groupnum}
\end{table*}

\end{document}